\documentclass{./styles/svproc}
\usepackage{url}
\usepackage{float}
\usepackage{graphicx}

\usepackage{cite}
\usepackage{amsmath,amssymb,amsfonts}
\usepackage{algorithmic}
\usepackage{textcomp}
\usepackage[table,xcdraw]{xcolor}
\usepackage{booktabs}
\usepackage{url}
\usepackage{makecell}
\usepackage{graphicx}
\usepackage{subcaption}

\begin{document}
\mainmatter              
\title{Comparative Analysis of CNN Performance in Keras, PyTorch and JAX on PathMNIST}
\titlerunning{CNN Performance in Keras, PyTorch and JAX on PathMNIST}  
%
 \author{Anida Nezović \and  Jalal Romano \and
 Nada Marić \and Medina Kapo \and Amila Akagić }

 \authorrunning{Nezović A., Romano J., Marić N., Kapo M., Akagić A.} 
%
 \tocauthor{Nezović A., Romano J., Marić N., Kapo M., Akagić A.}
 \institute{Faculty of Electrical Engineering, University of Sarajevo,  Sarajevo, Bosnia and Herzegovina,\\
 \email{\{anezovic1, jromano1, nmaric1, mkapo2, aakagic\}@etf.unsa.ba}
}

\maketitle              

\begin{abstract}
Deep learning has significantly advanced the field
of medical image classification, particularly with the adoption of
Convolutional Neural Networks (CNNs). Various deep learning
frameworks such as Keras, PyTorch and JAX offer unique
advantages in model development and deployment. However,
their comparative performance in medical imaging tasks remains
underexplored. This study presents a comprehensive analysis
of CNN implementations across these frameworks, using the
PathMNIST dataset as a benchmark. We evaluate training efficiency, classification accuracy and inference speed to assess their
suitability for real-world applications. Our findings highlight the
trade-offs between computational speed and model accuracy,
offering valuable insights for researchers and practitioners in
medical image analysis.
\end{abstract}

\section{Introduction}
The emergence of deep learning has revolutionized the field of medical image classification, offering high accuracy and robust feature extraction capabilities. Among various deep learning models, Convolutional Neural Networks (CNNs) have emerged as a dominant approach due to their ability to learn spatial hierarchies of features from image data\cite{cnnRad}. Several deep learning frameworks, including Keras, PyTorch and JAX, have gained popularity for their distinct computational paradigms and optimizations. Keras, built on TensorFlow, provides a high-level API for rapid prototyping and ease of use. PyTorch, known for its dynamic computational graph, offers flexibility and is widely adopted for research applications\cite{pytorchRad}. JAX, a relatively newer framework, leverages Just-In-Time (JIT) compilation and hardware acceleration to enhance computational efficiency, particularly for large-scale models\cite{jaxRad}.

Despite the extensive use of these frameworks, there remains a need for a comparative analysis that evaluates their strengths and weaknesses in medical image classification tasks. The PathMNIST dataset, serves as an ideal benchmark due to its structured nature and real-world relevance in digital pathology. This dataset contains images derived from histopathology slides of human colorectal cancer tissue, categorized into nine distinct tissue types.

This research focuses on evaluating the performance of Keras, PyTorch and JAX when applied to the PathMNIST dataset. By analyzing aspects such as training time and classification accuracy across these frameworks, this work aims to provide insights into their practical applicability in medical imaging tasks \cite{imgClassRad}.  The results of this study demonstrate that while all three frameworks are capable of achieving high classification accuracy, they differ significantly in terms of training efficiency. These findings underscore the importance of selecting a deep learning framework that aligns not only with performance goals but also with the specific constraints of a medical imaging project.

This paper is structured as follows: Section II reviews related work in medical image classification and framework comparisons. Section III describes the dataset and preprocessing steps. Section IV details the implementation of CNNs in each framework. Section V presents the results and evaluation metrics, followed by a discussion in Section VI. Finally, Section VII concludes the study and outlines potential future research directions.

\section{Related work}
Medical image classification has seen significant advancements with the integration of deep learning architectures. Traditional convolutional neural networks have been the foundation of many studies, but recent research has explored architectural improvements, optimization strategies and alternative deep learning models to enhance performance and computational efficiency.

One approach to improving CNN-based models is presented in~\cite{prviRad}, where a grammar-based method for generating CNN architectures is proposed. The authors leverage a context-free grammar combined with a multi-objective evolutionary algorithm to optimize the network design. The generated architectures balance performance and model simplicity. The proposed CNNs reduce parameter counts and training time compared to DenseNet169, maintaining comparable classification accuracy while improving efficiency.

Another study~\cite{cetvrtiRad} focuses on improving the robustness of medical image classification models against adversarial attacks and data limitations. The authors introduce Curvature Regularization, which stabilizes learning by controlling Hessian’s eigenvalues, reducing sensitivity to perturbations. The model outperforms traditional architectures such as ResNet-50. The results indicate that CURE-based models achieve higher classification accuracy with minimal performance degradation under adversarial conditions, making them suitable for medical diagnostics.

In terms of computational efficiency, a novel channel merging technique is introduced in~\cite{drugiRad} to address the excessive parameter count in deep CNNs. By replacing traditional concatenation operations with summation-based channel merging, the authors achieve a substantial reduction in floating point operations and memory consumption. Their experiments on CIFAR-10, CIFAR-100 and PathMNIST datasets reveal that the proposed method maintains accuracy while reducing parameter counts by over 60\%.

\begin{figure}
    \centering
   \includegraphics[width=2in]{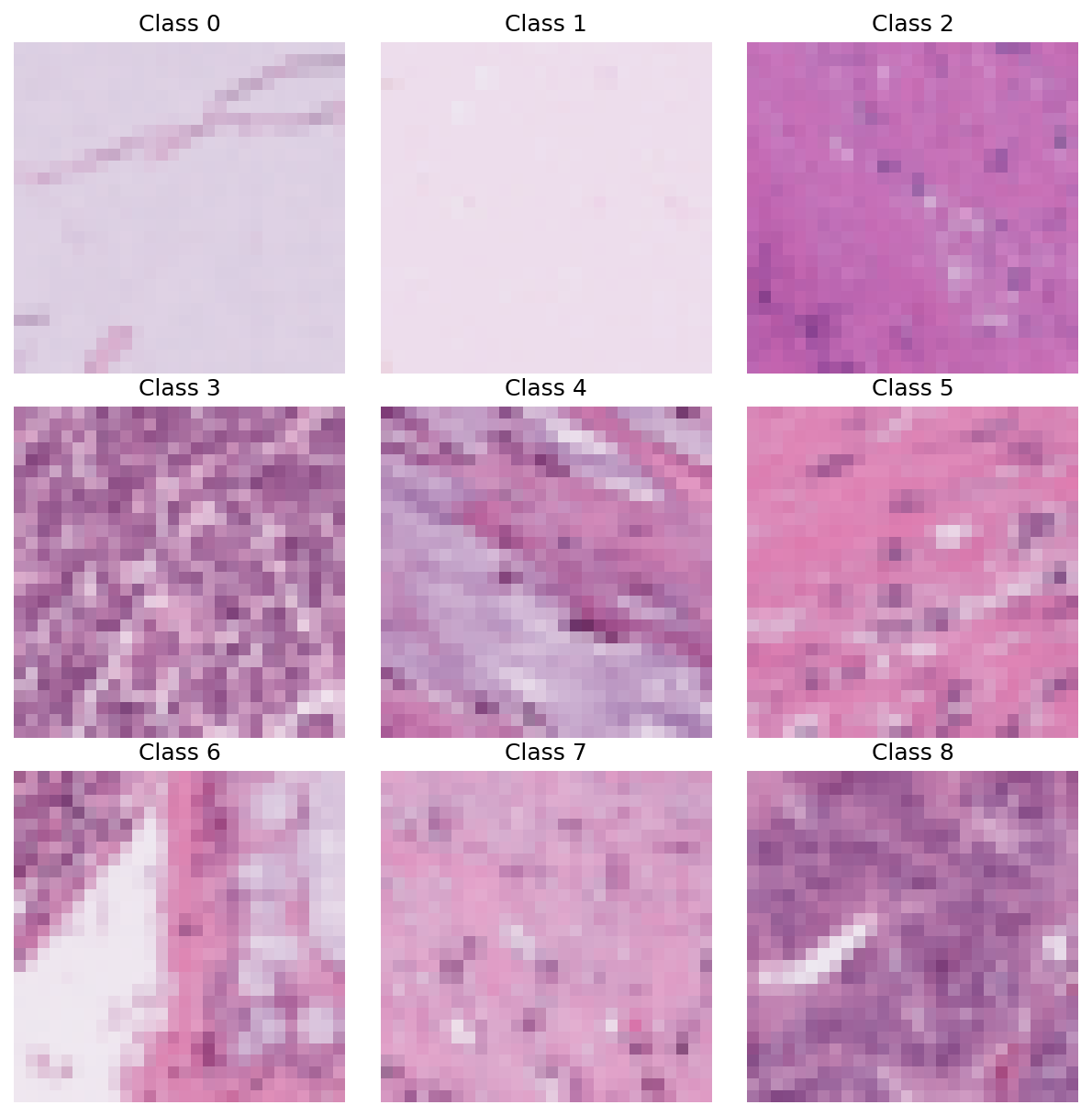}
    \caption{Various classes of PathMNIST dataset.}
    \label{fig:class_visualization}
\end{figure}

While CNNs remain dominant in image classification, alternative architectures such as Vision Transformers (ViTs) have gained traction. The study in~\cite{treciRad} explores the use of ViTs for medical image classification, highlighting their ability to capture long-range dependencies that CNNs struggle with due to locality bias. By leveraging a pre-trained ViT model fine-tuned on MedMNISTv2 datasets, the authors achieve superior performance compared to traditional CNNs. Their results demonstrate that ViTs surpass existing benchmarks, reinforcing their potential for medical image analysis. 

\section{Implementation}
In this section, the implementation details of the approach are described, including dataset preparation, preprocessing steps and model development. The dataset undergoes preprocessing to ensure high-quality input for training. Additionally, the workflow is outlined, detailing the models utilized and their role in the overall system.

\subsection{Dataset Overview}
The MedMNIST dataset collection is designed for lightweight biomedical image classification tasks, covering a wide range of medical imaging modalities. Among the datasets in MedMNIST, PathMNIST is specifically focused on histopathology images. It consists of nine classes representing different tissue types and pathological conditions shown in Fig. \ref{fig:class_visualization}, enabling multi-class classification tasks in medical imaging. The dataset was pre-divided into three subsets: training, validation and test sets. The training set consists of 89996 images, the validation set includes 10004 images and the test set contains 7180 images. Each image contains three color channels. To tailor the dataset to the specific objectives of this study, data preprocessing is applied. First, pixel values are normalized to the range [0,1] to standardize input values. Labels are then converted into one-hot encoding for compatibility with classification models\cite{mnistRad}. 

\begin{figure}
    \centering
    \includegraphics[width=0.65\textwidth]{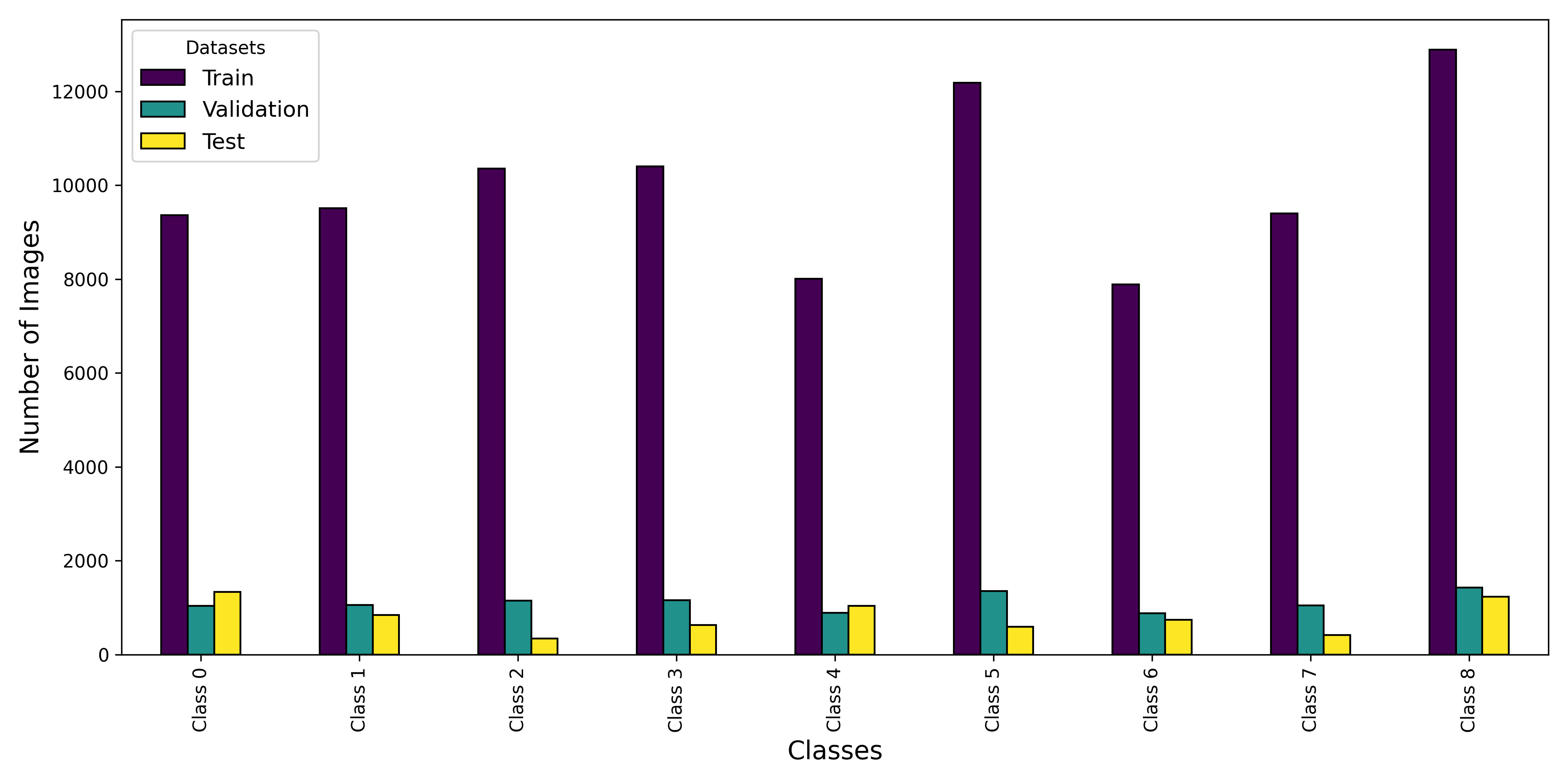} 
    \caption{Class distribution across the PathMNIST dataset.}
    \label{fig:class_distribution}
\end{figure}

The dataset distribution shows notable imbalances, particularly in the test set, where the proportions of some classes deviate significantly from the training and validation distributions (see Fig. \ref{fig:class_distribution}). Class 2 has only 4.72\% in the test set, while it had 11.51\% in training. Class 0 is overrepresented in the test set (18.64\%) relative to training (10.41\%). Class 1 is underrepresented in the test set (11.80\%) compared to training (10.57\%). Since some classes are underrepresented or overrepresented in the test set, model evaluation may be biased. Instead of excluding certain classes, class imbalance is addressed by incorporating weighted loss functions and penalty coefficients, giving higher importance to underrepresented classes during training. This ensures that the model learns effectively across all categories, mitigating the impact of imbalance on classification performance.

\subsection{Workflow}
The workflow of this study consists of three main stages: model implementation, evaluation and comparison. CNNs are implemented using three different deep learning frameworks: Keras, PyTorch and JAX. To ensure a fair comparison, a consistent model architecture is maintained across all frameworks. The architecture is determined after performing hyperparameter tuning, where different configurations of convolutional layers, activation functions, dropout rates and learning rates are tested \cite{hiperRad} (see Table \ref{tab:cnn_architecture}).

Once the models are trained on the PathMNIST dataset, an evaluation phase is conducted. The models are assessed using key performance metrics such as accuracy and training time to analyze their computational efficiency and classification performance. Finally, the results obtained from the evaluation are compared to highlight differences between frameworks in the context of medical image classification. The comparison provides insights into training efficiency, inference speed and overall suitability of each framework. All experiments were implemented in Jupyter Notebook using Python.

\section{Results}
This section presents the results obtained from training and evaluating the CNN on the PathMNIST dataset. The performance of the model is assessed using standard classification metrics. Additionally, the training process, computational efficiency and classification behavior are examined through performance plots and a confusion matrix.

\subsection{Evaluation approach}
Several metrics were used to assess the performance of the model. For each predicted sample, the number of true positives (TP), true negatives (TN), false positives (FP) and false negatives (FN) was computed \cite{metrikeRad}. In a multiclass setting, each class $i$ is treated in a one-vs-rest fashion, and the following counts are computed for each class:

\[
\mathrm{TP}_i = |\{\,x : y = i \ \wedge\ \hat y = i\}|,\quad
\mathrm{FP}_i = |\{\,x : y \neq i \ \wedge\ \hat y = i\}|, 
\]
\[
\mathrm{FN}_i = |\{\,x : y = i \ \wedge\ \hat y \neq i\}|,\quad
\mathrm{TN}_i = |\{\,x : y \neq i \ \wedge\ \hat y \neq i\}|.
\]

Here, $TP_i$ (true positives) counts samples correctly predicted as class $i$, $FP_i$ (false positives) counts samples from other classes mistakenly predicted as $i$, $FN_i$ (false negatives) counts samples of class $i$ predicted as a different class, and $TN_i$ (true negatives) counts all remaining correctly classified samples. These values form the basis for computing precision, recall, F1-score and accuracy for each class.

\subsubsection{Accuracy}
Measures the proportion of correctly classified samples.
\[
  \mathrm{Accuracy}
  = \frac{\sum_{i=1}^K \mathrm{TP}_i}
         {\sum_{i=1}^K (\mathrm{TP}_i + \mathrm{FP}_i + \mathrm{FN}_i)}
  = \frac{\sum_{i=1}^K \mathrm{TP}_i}{N},
\]
where \(N\) is the total number of samples.

\subsubsection{Precision}
Assesses the fraction of correctly predicted positive instances among all predicted positives.
\begin{equation}
  \mathrm{Precision}_i
  = \frac{\mathrm{TP}_i}{\mathrm{TP}_i + \mathrm{FP}_i}
\end{equation}

\subsubsection{Recall (Sensitivity)}
Determines the proportion of actual positive samples that were correctly identified.
\begin{equation}
  \mathrm{Recall}_i
  = \frac{\mathrm{TP}_i}{\mathrm{TP}_i + \mathrm{FN}_i}
\end{equation}

\subsubsection{F1-score}
Provides a balance between precision and recall.
\begin{equation}
  \mathrm{F1-score}_i
  = 2 \times \frac{\mathrm{Precision}_i \times \mathrm{Recall}_i}
                     {\mathrm{Precision}_i + \mathrm{Recall}_i}
\end{equation}

These metrics provide a comprehensive evaluation of the classification model by considering both overall performance and class-wise misclassification tendencies.

\subsection{Result analysis}
The CNN implemented in Keras was trained for 20 epochs per run, across 10 independent runs to ensure result consistency. It achieved an average training accuracy of 96.04\% and an average validation accuracy of 95.58\%. The average test accuracy across iterations was 90\%. The total training time averaged 642.59 seconds, while the inference phase, conducted on 7180 test samples, resulted in an average inference time of 2.3036 seconds. 

\begin{figure}[ht]
    \centering
    \begin{subfigure}[b]{0.3\textwidth}
        \includegraphics[width=\textwidth]{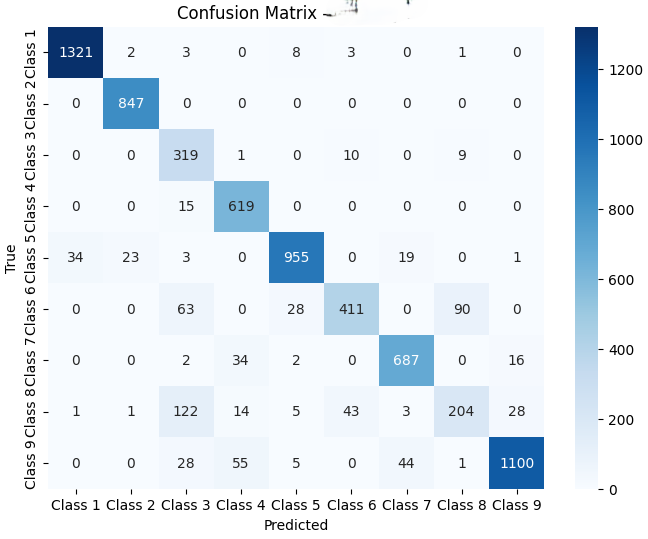}
        \caption{Keras}
        \label{fig:keras_conf_matrix}
    \end{subfigure}
    \hfill
    \begin{subfigure}[b]{0.3\textwidth}
        \includegraphics[width=\textwidth]{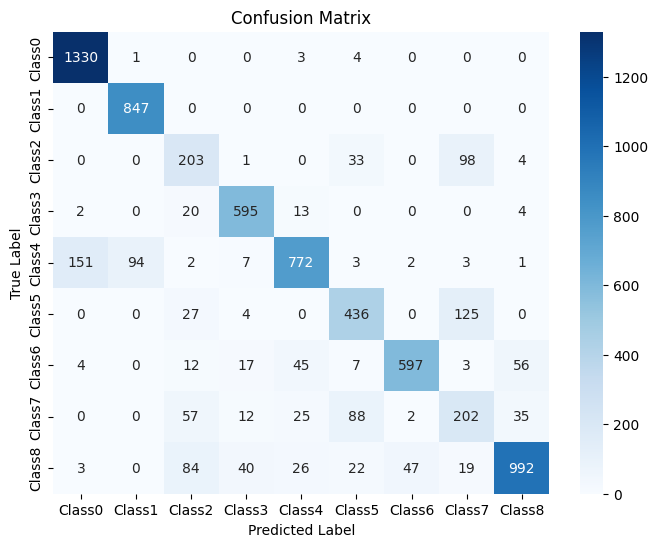}
        \caption{JAX}
        \label{fig:jax_conf_matrix}
    \end{subfigure}
    \hfill
    \begin{subfigure}[b]{0.3\textwidth}
        \includegraphics[width=\textwidth]{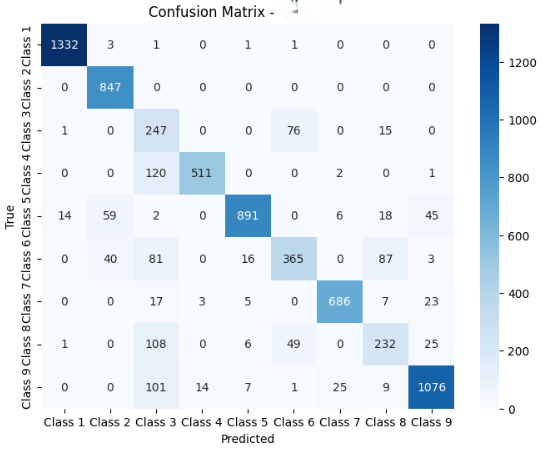}
        \caption{PyTorch}
        \label{fig:pytorch_conf_matrix}
    \end{subfigure}
    
    \caption{Confusion matrices produced by models implemented in (a) Keras, (b) JAX, and (c) PyTorch for the PathMNIST dataset.}
    \label{fig:conf_matrices}
\end{figure}

The classification performance of the Keras model is further analyzed using the confusion matrix, as shown in Figure~\ref{fig:conf_matrices}\subref{fig:keras_conf_matrix}. The highest-performing classes include Class 0, 1 and 9, with F1-scores of 0.98, 0.98 and 0.92, respectively. Class 0 demonstrates strong classification performance with high precision (0.97) and recall (0.99), indicating minimal misclassification.

\begin{table}[H]
    \centering
    \caption{\textnormal{Architecture and hyperparameters of the selected CNN}}
    \label{tab:cnn_architecture}
    \rowcolors{2}{blue!10!white}{white}  
    \begin{tabular}{l c c l}
        \toprule
        \textbf{Layer Type} & \textbf{Kernel / Units} & \textbf{Activation} & \textbf{Hyperparameters} \\
        \midrule
        Conv2D  & $64 \times (3 \times 3)$  & ReLU  & Padding: Same \\
        Conv2D  & $64 \times (3 \times 3)$  & ReLU  & Padding: Same \\
        MaxPooling2D & $2 \times 2$  & -  & Stride: 2 \\
        Dropout  & -  & -  & Rate: 0.3 \\
        \midrule
        Conv2D  & $128 \times (3 \times 3)$  & ReLU  & Padding: Same \\
        Conv2D  & $128 \times (3 \times 3)$  & ReLU  & Padding: Same \\
        MaxPooling2D & $2 \times 2$  & -  & Stride: 2 \\
        Dropout  & -  & -  & Rate: 0.3 \\
        \midrule
        Conv2D  & $128 \times (3 \times 3)$  & ReLU  & Padding: Same \\
        Conv2D  & $128 \times (3 \times 3)$  & ReLU  & Padding: Same \\
        MaxPooling2D & $2 \times 2$  & -  & Stride: 2 \\
        Dropout  & -  & -  & Rate: 0.3 \\
        \midrule
        Flatten & -  & -  & - \\
        Dense  & 256  & ReLU  & - \\
        Dropout  & -  & -  & Rate: 0.3 \\
        Dense  & 128  & ReLU  & - \\
        Dropout  & -  & -  & Rate: 0.3 \\
        Dense  & $5$   & Softmax  & - \\
        \midrule
        \multicolumn{4}{c}{\textbf{Training Hyperparameters}} \\
        \midrule
        \textbf{Optimizer}  & \multicolumn{3}{c}{Adam} \\
        \textbf{Learning Rate}  & \multicolumn{3}{c}{$0.00027092$} \\
        \textbf{Loss Function}  & \multicolumn{3}{c}{Categorical Crossentropy} \\
        \textbf{Batch Size}  & \multicolumn{3}{c}{32} \\
        \textbf{Epochs}  & \multicolumn{3}{c}{20} \\
        \bottomrule
    \end{tabular}
\end{table}
 Similarly, Class 1 achieves a perfect recall (1.00). However, certain classes exhibit lower classification performance. Class 7 has the lowest F1-score (0.56), primarily due to a lower recall of 0.48, suggesting that many of its instances are misclassified. This could be attributed to visual or structural similarities with other classes, particularly Class 5 and Class 9. Additionally, Class 3 has a lower precision (0.82) but maintains a high recall (0.98).


The neural network was also implemented using JAX with identical hyperparameter values and trained for 20 epochs across 10 independent runs. The average training and validation accuracies achieved were 77.22\% and 80.12\%, respectively. When it comes to the total training time, the JAX implementation outperformed the Keras implementation, completing training in an average of 343.14 seconds. The model was again evaluated on the test dataset, achieving an average accuracy of 76.54\%. For inference JAX again surpassed Keras, processing the same 7180 test samples in an average of just 0.2795 seconds, yielding an average inference time of 0.03893 milliseconds per sample. The training process of the JAX model doesn't monotonically improve over different epochs.\\ 

The confusion matrix for the results obtained with the JAX model gives us more insight on its performance, as shown in Figure~\ref{fig:conf_matrices}\subref{fig:jax_conf_matrix}. JAX also obtained high precision, recall and F1-scores for some classes. The highes per-class accuracy are achieved for Class 1 and Class 3, with accuracies of 1.0 and 0.99, respectively. The lowest precision of 0.11 is got for Class 5. In this case, Class 5 shows the most false positive predictions, having the recall of 0.12. Macro average F1 score 0.62 and weighted average F1 score 0.71, show us that the JAX model does not have balanced classification across all classes.


The neural network was also implemented in PyTorch using the same architecture and hyperparameters as the previous implementations. The model was trained across 10 independent runs, each lasting 20 epochs. It achieved an average training accuracy of 95.64\%, an average validation accuracy of 95.17\%, and an average test accuracy of 86.48\%. The average training time for PyTorch was 375.31 seconds, making it significantly faster than Keras but slower than JAX in terms of training efficiency. The inference phase, conducted on the same 7180 test samples, took an average of 0.4980 seconds, resulting in an average inference time per sample of approximately 0.120 milliseconds. In terms of performance, Keras achieved the highest accuracy, while PyTorch positioned itself between Keras and JAX, offering a balance between speed and classification performance.\\
The confusion matrix for the PyTorch implementation, illustrated in Figure~\ref{fig:conf_matrices}\subref{fig:pytorch_conf_matrix}, provides further insight into its classification performance. The model demonstrated exceptionally high precision in classifying Classes 1 and 2, with precision values of 0.99, indicating that the model rarely misclassified other classes as these. Class 2 also achieved the highest recall (1.00). However, Class 8 had the lowest recall (0.60), suggesting that many instances belonging to this category were misclassified as other classes. Class 3 exhibited the highest number of false positives (228), meaning that the model frequently predicted this class incorrectly. Additionally, Class 9, despite having a relatively high precision (0.98), had a recall of 0.88. These results suggest that while PyTorch achieved competitive accuracy and efficiency, JAX was superior in inference speed, and Keras achieved the highest accuracy on the test set.



\section{Discussion}
The experimental results reveal significant differences in training and inference efficiency among Keras, PyTorch, and JAX (see Table~\ref{tab:framework_comparison}). When it comes to the total training time required, JAX demonstrates the highest computational efficiency, completing training in 343.14s, which is 46.6\% faster than Keras (642.59s) and 8.57\% faster than PyTorch  (375.31s). PyTorch, while slightly slower than JAX, still outperforms Keras by 41.59\%, indicating a more optimized execution pipeline.

Inference speed on the 7180 samples, further highlights JAX’s computational advantage. JAX achieves the lowest total inference time of 0.2795s, making it 8.24 times faster than Keras (2.3036s) and 1.78 times faster than PyTorch (0.4980s). These results suggest that JAX’s just-in-time (JIT) compilation and optimized execution graph significantly enhance computational efficiency. Pytorch again outperforms Keras, reducing the inference time 4.62 times.

\begin{table}
    \centering
    \begin{tabular}{lcccc}
        \hline
        \textbf{Framework} & \makecell{\textbf{Training} \\ \textbf{time (s)}} & \makecell{\textbf{Inference} \\ \textbf{time (s)}} & \makecell{\textbf{Accuracy}} \\
        \hline
        Keras  & \makecell{642.59}  & \makecell{2.304}  & \makecell{0.9} \\
        PyTorch & \makecell{375.31} & \makecell{0.498} & \makecell{0.86}   \\
        JAX    & \makecell{343.14} & \makecell{0.279} & \makecell{0.76}    \\
        \hline
    \end{tabular}
    \caption{Comparison of CNN Performance Across Frameworks}
    \label{tab:framework_comparison}
\end{table}

Regarding the model's classification performance, not all three frameworks achieve high accuracy. There are noticeable  differences in precision, recall, and F1-scores for individual classes, which highlight 
the trade-offs between these frameworks. 

The Keras model achieves the highest overall accuracy of 0.90, followed by the PyTorch model with 0.86, and the JAX model with 0.76. 
Per-class performance varies across these frameworks as well.

\begin{itemize}

\item High-Performing Classes: 
All frameworks excel in Class 0 (near perfect precision/recall). For Class 1, Keras and JAX lead (F1: 0.98) and PyTorch follows closely (F1: 0.97) with perfect recall across all three frameworks (1.00). 

\item Precision-Recall Trade-offs:
Class 8 shows high precision but lower recall (Keras: 0.89, JAX: 0.88, PyTorch: 0.62), indicating that the model is very accurate in its positive predictions but fails to identify a significant portion of actual positive instances. For Class 3, PyTorch achieves the best balance, while Keras and JAX with precisions 0.82 and 0.53 respectively, prioritize recall (0.98 and 0.99) at the cost of false positives.

\item Underperforming Classes:
Performance drops sharply in Class 7: Keras (F1: 0.56) and JAX (F1: 0.35) struggle, and PyTorch achieves a lower recall of 0.6, indicating that the model does not perform well with this class.
\end{itemize}

Although JAX consistently outperforms both PyTorch and Keras in terms of training and inference speed, its classification accuracy remains notably lower. This discrepancy may be attributed to several technical factors inherent to JAX’s architecture and design. Unlike Keras and PyTorch, JAX relies on a functional programming model and uses Just-In-Time (JIT) compilation via the XLA compiler to transform Python functions into highly optimized machine code. This results in faster execution but may introduce subtle differences in numerical precision or gradient calculations due to operation reordering or aggressive optimizations. XLA-accelerated kernels can produce different floating-point results than TensorFlow or PyTorch which can cause numerical error accumulation.
After evaluating the models' performances, it is evident that each framework is suited for different use cases.
Considering that this is a medical classification task, minimizing false negatives is critical, making Keras a strong choice due to its high recall. PyTorch is a strong contender as well, as it offers a balanced approach, performing well across most classes.
On the other hand, JAX stands out as the fastest framework, but comes with a significant cost to accuracy, making it better suited for large-scale or real-time applications where speed is a priority.


\section{Conclusion}
Experimental results demonstrate that while all three frameworks achieve comparable accuracy levels, JAX exhibits superior computational efficiency, as it requires less computational time for training and inference. Specifically, JAX reduced training time by 46.6\% compared to Keras and 8.57\% compared to PyTorch. PyTorch, on the other hand, balanced flexibility and performance, whereas Keras provided ease of use but at the cost of increased computational overhead. Our findings highlight the trade-offs between computational speed and model accuracy across these deep learning frameworks.

A similar comparative study could be conducted on each dataset within the MedMNIST collection to verify whether the observed framework trade-offs hold across diverse imaging modalities and class distributions. For example, BreastMNIST is a binary classification task (benign versus malignant histology) with highly skewed class distributions, which will reveal how each framework handles severe data imbalance and a limited number of minority-class samples. PneumoniaMNIST features chest X-rays in two classes, with higher resolution and greater variance in patient positioning, thereby placing pressure on memory management and preprocessing speed. OrganMNIST contains eleven classes of two-dimensional cross-sections from CT and MRI scans, where multiscale anatomical structures and projection artifacts demand a model’s capacity to capture both global context and local detail. Finally, DermaMNIST offers seven classes of color skin lesion images marked by variation in lighting, color and real-world artifacts, making it an ideal benchmark for evaluating end-to-end augmentation, transfer-learning approaches, and fine-tuning capabilities.

In future research it would be insightful to explore the impact of other hardware accelerators and data augmentation for underrepresented classes in the PathMNIST dataset, to further assess the scalability and usability of these frameworks. For additional frameworks, boilerplate code could be drastically reduced by adopting PyTorch Lightning or FastAI and multi-GPU training workflows could be streamlined. Although less widely adopted, MindSpore or MXNet Gluon claim efficient distributed training and lower latency inference. Including them in a cross-dataset study will help determine whether their performance claims hold up in a variety of medical-imaging contexts. By benchmarking TensorFlow 2.x + XLA during the training phase and TensorFlow Lite during inference across all MedMNIST datasets, one can quantify the end-to-end speed gains, memory savings and any accuracy trade-offs.

Additionally, extending the analysis to other deep learning architectures could provide additional insights into these frameworks' performances in medical classification applications. Vision Transformers (ViTs) can be fine-tuned on a dataset to determine whether long-range dependency modeling yields higher accuracy on tasks with large backgrounds or low-contrast targets. Hybrid architectures, in which CNN front-ends are combined with Transformer blocks, can be compared against purely convolutional networks to evaluate whether the additional modeling capacity justifies the increased computational cost.

\section{Declaration of conflicts of interest}
The authors declare that there are no conflicts of interest related to this paper.


%
%

\end{document}